\documentclass[10pt]{article}

\usepackage{geometry} 
\usepackage{graphicx} 

\usepackage{booktabs} 
\usepackage{array} 
\usepackage{paralist} 
\usepackage{verbatim} 
\usepackage{subfig} 

\usepackage{fancyhdr} 
\usepackage{sectsty}

\usepackage[nottoc,notlof,notlot]{tocbibind} 
\usepackage[titles,subfigure]{tocloft} 

\title{Genetic cellular neural networks for generating three-dimensional geometry}
\author{Hugo Martay\footnote{h.martay1@gmail.com}}
\date{2015-03-19} 

\begin{document}
\maketitle

\begin{abstract}
There are a number of ways to procedurally generate interesting three-dimensional shapes, and a method where a cellular neural network is combined with a mesh growth algorithm is presented here. The aim is to create a shape from a genetic code in such a way that a crude search can find interesting shapes.

Identical neural networks are placed at each vertex of a mesh which can communicate with neural networks on neighboring vertices. The output of the neural networks determine how the mesh grows, allowing interesting shapes to be produced emergently, mimicking some of the complexity of biological organism development. Since the neural networks' parameters can be freely mutated, the approach is amenable for use in a genetic algorithm.
\end{abstract}

\section{Overview}

The most obvious way to generate a three-dimensional mesh in a mutateable way would be to simply take a representation of the shape, and directly mutate it. If the shape was the level set of a sum of spherical harmonics, then you could just mutate the proportions of each spherical harmonic, and the shape would change correspondingly. In a shape represented by a mesh, the mesh vertices could be mutated directly.

In biology, the way that morphologies can be mutated seems richer than in either of these examples. For instance, in both of the examples above, a child organism would be unlikely to be just a scaled version of its parent, because too many mutations would have to coincide. It would be unlikely to find left-right symmetry evolving in either of the above methods unless the morphology was explicitly constrained. In nature, the link between morphology and the organism's genome is much more complicated than for the examples above.

Modeling the chemical processes behind the development of an organism is an active field, which is described in detail in \textit{On growth, form and computers}\cite{kumar2003growth}. A widely used model that describes organism development has been presented by Kumar and Bentley~\cite{kumar_bentley2003}. 

In this work, the same philosophy is adopted, that an emergent system needs to be parameterised by a genome, and the morphology needs to be a result of the system's dynamics. However, the emergent system used here is a network of identical neural networks, or cellular neural network. These were described by Chua and Yang~\cite{chua1988cellularT} and a well-known review of work on cellular neural networks was written by Cimagalli and Balsi~\cite{cimagalli1993cellular}. 

A paper by Wilfried Elmenreich and Istv\'{a}n Feh\'{e}rv\'{a}ri~\cite{elmenreich2011evolving} uses a cellular neural network to reconstruct images, and the architecture of their cellular neural network appears similar to the method in this paper. Here, though, the output of the network is information that is used to grow a mesh, and the top-level topology of the cellular neural network here has to be flexible enough to allow for differing numbers of neighbours for each cell.

Cellular neural networks are capable of simulating a large variety of systems, and have been demonstrated to be able to model Conway's game of life \cite{Gomez-Ramirez2007}, which is known to be Turing complete, and so it is at least plausible that they could generate complicated structured patterns that resemble biological morphologies. 

\section{Method}

The calculation presented here takes place on a network of vertices. There are a certain number of discrete timesteps.
Each vertex, $i$, at each time, $t$, has a real-valued output vector, $\mathbf{w}_{i,t}$.

Each vertex has a number of neighboring vertices. 

Each vertex, at each timestep has an input vector, $\mathbf{v}_{i,t}$, such that $\mathbf{v}_{i,t}$ is a function of neighboring vertices' outputs in the previous timestep:

$$ \mathbf{v}_{i,t} = \mathbf{F}^i(\mathbf{w}_{a,t-1}, \mathbf{w}_{b,t-1}, \ldots),$$

\noindent where $a$,$b$,$\ldots$ is the set of neighbours of vertex $i$. The function that maps from neighboring outputs to inputs, $\mathbf{F}^i$, is given the superscript, $i$, to denote that it can vary from vertex to vertex. This is simply to allow for slightly different processing when the vertex might have different numbers of neighbours or have a slightly different geometry.

The mapping from input, $\mathbf{v}_{i,t}$, to output, $\mathbf{w}_{i,t}$ is calculated using a feed-forward neural network with a sigmoid activation function.
 
\subsection{Neural Network}

The neural network is described in C-like pseudocode:

\begin{verbatim}
double[] Evaluate(double[] input)
{
    for (int j = 0; j < input.Length; j++) neuron[0][j].value = input[j];
    for (int i = 1; i < NLayers; i++) // The zero-th layer is skipped.
    {
        for (int j = 0; j < number of neurons in layer i; j++)
        {
            double a = -neuron[i][j].threshold;
            for (int k = 0; k < number of neurons in layer (i-1); k++)
                a += (neuron[i - 1][k].value - 0.5) * neuron[i][j].weights[k];
            neuron[i][j].value = 1.0 / (1.0 + Exp(-a));
        }
    }
    for (int j = 0; j < number of neurons in final layer; j++) 
        output[j] = neuron[last][j].value;
    return output;
}
\end{verbatim}

The neural network is parameterised by each neuron's weights vector and threshold. These form the mesh's genetic code - any mutation or crossover or other operation on the mesh's genetic code simply varies these weights and thresholds. 

\subsection{Vertex Network}

The vertex network here is a three dimensional mesh - consisting of vertices with a three dimensional position, and faces with three vertices. Each vertex's neighbours are any vertex with which it shares a face. The input function, $\mathbf{F}$, that gives each vertex's input vector as a function of its neighbours' output vectors, does the following:

\begin{itemize}
\item If the output vector is length $N$, then the input vector is length $3N$: Each output number becomes three input numbers regardless of the mesh topology.

\item The first input number to this vertex is its own output from the previous timestep.

\item The second input number is the average output from all its neighbours from the previous timestep.

\item The third input number is a measure of the dispersion of its neighbours.

\item Some inputs are reserved for things like the orientation of the vertex or its adjacent faces.

\end{itemize}

In this way, each vertex can communicate with its neighbours, but in a symmetry preserving way. 

\begin{figure}[h]
    \centering
    \includegraphics[width=0.6\textwidth]{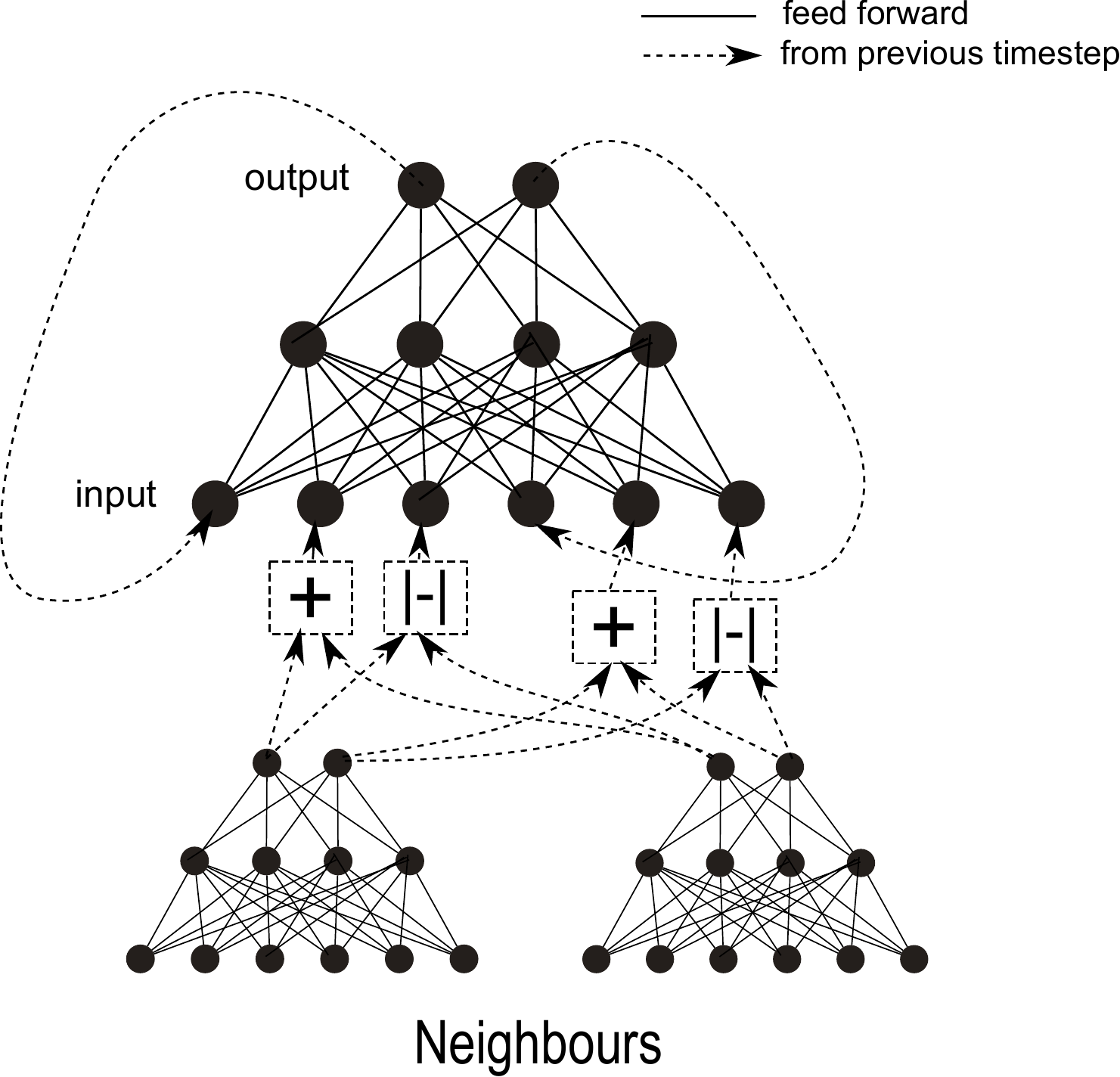}
    \caption{In this figure, each vertex is a neural network with six inputs and two outputs. In each timestep, the inputs are gathered from the output of the same neural network and the outputs of its neighboring neural networks, from the previous timestep. }
    \label{archnet}
\end{figure}

The architecture of this network and how it relates to the neural networks on each vertex is shown in figure~\ref{archnet}

\subsection{Mesh growth}
The timestep has one final component: The mesh updates according to the vertex output vector. Each vertex has a 
normalised three-dimensional vector, $\mathbf{h}_i$, describing the direction that it's position, $\mathbf{x}_i$ can grow. It then grows according to the following:

$$\mathbf{x}_{i, t+1} - \mathbf{x}_{i, t} = k \mathbf{h}_i \hbox{max}\left\lbrace 0, w_{i,t,0}-\frac{1}{2}\right\rbrace,$$

\noindent where $k$ is a normalization factor so that the mesh as a whole can only grow so fast, and $w_{i,t,0}$ is the 
zero-th element of the output of vertex $i$ at this time.

The mesh then checks to see if any face has an area that is above a threshold (which in some cases can be altered by a different one of the output elements of its vertices), and if so, places a new vertex in the middle of the face, and replaces itself with three new faces that integrate the new central vertex. The growing direction, $\mathbf{h}$, for the new vertex depends on both the normal of the original face, and a weighted sum of the growing directions of the three parent vertices, the weighting being determined by a vertex output element.

Finally, if any faces share two vertices, adjacent faces are checked to see if it they would be improved by switching the common edge: faces $ABD$ and $BCD$
might be rearranged as $ABC$, $CDA$, depending on their relative orientation, the length $BC$ compared to $AD$, and whether vertices $B$ or $D$ have already got too few adjacent faces (since this operation would reduce that number).

\subsection{Method summary}

Start with a  simple three-dimensional mesh with vertices and triangular faces. Assign each vertex an output vector of length $N$, and a growing direction, $\mathbf{h}$.

For each timestep, 
\begin{itemize}
\item Calculate a length $3N$ input vector for each vertex based on the outputs of its neighbours.
\item Calculate each vertex's output vector using a neural network --- each vertex has an identical neural network to the others.
\item Update the mesh according to the vertex output: Move the vertices, check to see if any new vertices should be added, and adjust the mesh accordingly, and consider switching a few edges.
\end{itemize}

In the examples shown below, there are fifteen outputs, forty-five inputs and thirty neurons in the hidden layer. Of the forty-five inputs, four are overridden with the vertex growth direction (three inputs), the height of the vertex (one input). Three outputs are used to guide mesh growth: one moves the vertex along its growth direction, one influences the area required for a face split, and the last influences the growth direction of any vertices that are placed as a result of a face splitting that this vertex is part of.

\subsection{Genetic algorithm}
It is outside the scope of this document to discuss how to implement a genetic algorithm, but the basic idea is that you 
have a population of genomes, each specifying the free parameters of a neural network that generates the 3d mesh. The population gradually replaces worse genomes with better ones, and generates new genomes from old ones, allowing for mutation and optionally crossover.

\section{Results}
\begin{figure}[h]
    \centering
    \includegraphics[width=1.0\textwidth]{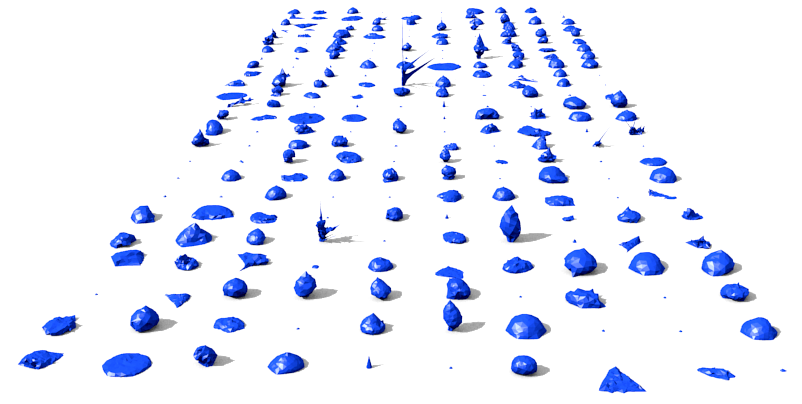}
    \caption{This shows two hundred and fifty meshes from randomly chosen genomes: No selection has been made. Almost all the meshes are distinct from one another, and several interesting shapes have emerged. This perhaps suggests that finding interesting shapes that maximise some heuristic using a search method such as a genetic algorithm should be easy.}
    \label{UnSelected}
\end{figure}

If the genomes are chosen completely randomly, with each weight or threshold chosen to be evenly distributed between -2 and +2, and run the algorithm for 200 timesteps, then the shapes generated look like those shown in figure~\ref{UnSelected}. Note that even without selection, the shapes are quite diverse and already slightly interesting. Not many genomes have simply failed to produce any mesh other than the initial.

\begin{figure}[h]
    \centering
    \includegraphics[width=0.5\textwidth]{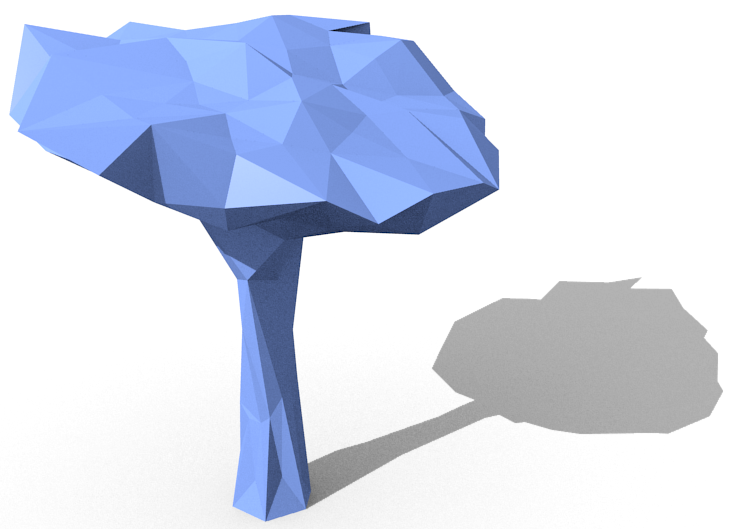}
    \caption{If a heuristic that mimics the selection pressure of trees in a forest is used to select meshes, you get a shape that resembles a tree. This demonstrates how the approach described here can be used in a genetic algorithm. }
    \label{over10area}
\end{figure}

An example is given here to demonstrate that this is suitable for use in genetic algorithms, where meshes are chosen to maximise the heuristic

$$\frac{\int \frac{1}{1+e^{10-h(x,y)}} dx dy}{1000 + \hbox{vertex count}},$$

\noindent where $h(x,y)$ is the maximum z-height of the mesh at that x,y position. This has been chosen to roughly mimic the selection pressure on trees --- the larger the horizontal surface area, the more light the organism will receive, but only if high enough to escape the shade of competitors. This heuristic encourages the formation of a canopy above ten units of height, and the mesh will necessarily have a trunk in order to reach that height. An example after several generations of selection is show in figure~\ref{over10area}. It is apparent that the shape look somewhat like a tree.

\section{Discussion}

A fairly simple algorithm is presented that can generate interesting shapes according to a genetic code that is suitable for use in a genetic algorithm. An example heuristic was given which demonstrated its suitability for use in a genetic algorithm.

The neural networks here are meant to be an analogy to the process that forms morphology in nature: The outputs 
of the neural network are meant to be analogous to the state of the cells in this region of the organism, including any chemical markers or hormones that might influence the organism's local growth. Neural network were chosen to be the equivalent process in these simulations because the implementation is simple.

This technique might find applications in reconstructing asteroid shapes from light curves, or in biomedical imaging, or in computer graphics. 

\section{Meta}

The method outlined here was developed in a non-academic setting, and published because it appears to be novel. However, the author acknowledges that there may be relevant papers that should have been cited but were neglected. Any comments or suggestions would be gratefully received.

The code used in this document is available as part of the arxiv submission: Its inclusion clearly makes the work easier to replicate and use, since any ambiguity in the text can be resolved. 
\bibliographystyle{plain}

\end{document}